\documentclass[10pt,twocolumn,letterpaper]{article}

\usepackage{cvpr}              %

\makeatletter
\@namedef{ver@everyshi.sty}{}
\makeatother

\usepackage{graphicx}
\usepackage{amsmath}
\usepackage{amssymb}
\usepackage{booktabs}
\usepackage{microtype}
\usepackage{xspace}
\usepackage{amsmath,amsfonts,amssymb,amsthm}
\usepackage[dvipsnames]{xcolor}
\usepackage{diagbox}
\usepackage{pifont}
\usepackage{svg}
\usepackage{amsmath}
\usepackage[numbers]{natbib}
\usepackage{enumitem}
\usepackage{physics}
\usepackage{mathtools}
\usepackage{algorithm}
\usepackage{algorithmic}
\usepackage{gensymb}
\usepackage{wrapfig}
\usepackage{subcaption}
\usepackage{multirow}
\usepackage{footmisc}
\usepackage{empheq}
\usepackage{cases}
\usepackage{adjustbox}
\usepackage{array}
\usepackage{tabu}
\usepackage{flushend}
\usepackage{multirow}
\usepackage{bm}
\usepackage{color, colortbl}

\usepackage[pagebackref,breaklinks,colorlinks]{hyperref}

\usepackage[capitalize]{cleveref}
\crefname{section}{Sec.}{Secs.}
\Crefname{section}{Section}{Sections}
\Crefname{table}{Table}{Tables}
\crefname{table}{Tab.}{Tabs.}

\DeclareMathOperator*{\argmax}{arg\,max}
\DeclareMathOperator*{\argmin}{arg\,min}

\theoremstyle{plain}

\theoremstyle{definition}

\theoremstyle{remark}

\def\cvprPaperID{233} %
\def\confName{CVPR}
\def\confYear{2023}

\begin{document}

\makeatletter
\DeclareRobustCommand\onedot{\futurelet\@let@token\@onedot}
\def\@onedot{\ifx\@let@token.\else.\null\fi\xspace}

\def\eg{\emph{e.g.}\xspace} \def\Eg{\emph{E.g}\onedot}
\def\ie{\emph{i.e.}\xspace} \def\Ie{\emph{I.e}\onedot}
\def\vs{\emph{vs.}\xspace}
\def\cf{\emph{c.f}\onedot} \def\Cf{\emph{C.f}\onedot}
\def\etc{\emph{etc}\onedot} \def\vs{\emph{vs}\onedot}
\def\wrt{w.r.t\onedot} \def\dof{d.o.f\onedot}
\def\etal{\emph{et al}\onedot}
\def\viz{\emph{viz}\onedot}
\makeatother

\newcommand{\cmark}{\ding{51}}%
\newcommand{\xmark}{\ding{55}}%

\newcommand{\bbR}{\mathbb{R}}
\newcommand{\bfx}{\mathbf{x}}
\newcommand{\bfh}{\mathbf{h}}
\newcommand{\bfz}{\mathbf{z}}
\newcommand{\bfZ}{\mathbf{Z}}
\newcommand{\tbfz}{\widetilde{\mathbf{z}}}
\newcommand{\tbfZ}{\widetilde{\mathbf{Z}}}
\newcommand{\bfu}{\mathbf{u}}
\newcommand{\bfU}{\mathbf{U}}
\newcommand{\tbfu}{\widetilde{\mathbf{u}}}
\newcommand{\tbfU}{\widetilde{\mathbf{U}}}
\newcommand{\bfQ}{\mathbf{Q}}
\newcommand{\bfK}{\mathbf{K}}
\newcommand{\bfV}{\mathbf{V}}
\newcommand{\bfP}{\mathbf{P}}
\newcommand{\bfJ}{\mathbf{J}}

\newcommand{\model}{{AbSViT}\xspace} 
\newcommand\minisection[1]{\vspace{1.3mm}\noindent \textbf{#1}}

\definecolor{Gray}{gray}{0.9}

\title{Top-Down Visual Attention from Analysis by Synthesis}

\author{Baifeng Shi\\
UC Berkeley\\
\and
Trevor Darrell\\
UC Berkeley\\
\and
Xin Wang\\
Microsoft Research\\
}
\maketitle
\begin{abstract}

Current attention algorithms (e.g., self-attention) are stimulus-driven and highlight all the salient objects in an image.
However, intelligent agents like humans often guide their attention based on the high-level task at hand, focusing only on task-related objects. 
This ability of task-guided top-down attention provides task-adaptive representation and helps the model generalize to various tasks. 
In this paper, we consider top-down attention from a classic Analysis-by-Synthesis (AbS) perspective of vision. Prior work indicates a functional equivalence between visual attention and sparse reconstruction; we show that an AbS visual system that optimizes a similar sparse reconstruction objective modulated by a goal-directed top-down signal naturally simulates top-down attention. We further propose Analysis-by-Synthesis Vision Transformer (\model), which is a top-down modulated ViT model that variationally approximates AbS, and achieves controllable top-down attention. For real-world applications, \model consistently improves over baselines on Vision-Language tasks such as VQA and zero-shot retrieval where language guides the top-down attention. \model can also serve as a general backbone, improving performance on classification, semantic segmentation, and model robustness.
Project page: \url{https://sites.google.com/view/absvit}.
   
\end{abstract}

\section{Introduction}
\label{sec:intro}

\begin{figure}[t]
\begin{center}
\centerline{\includegraphics[width=1\columnwidth]{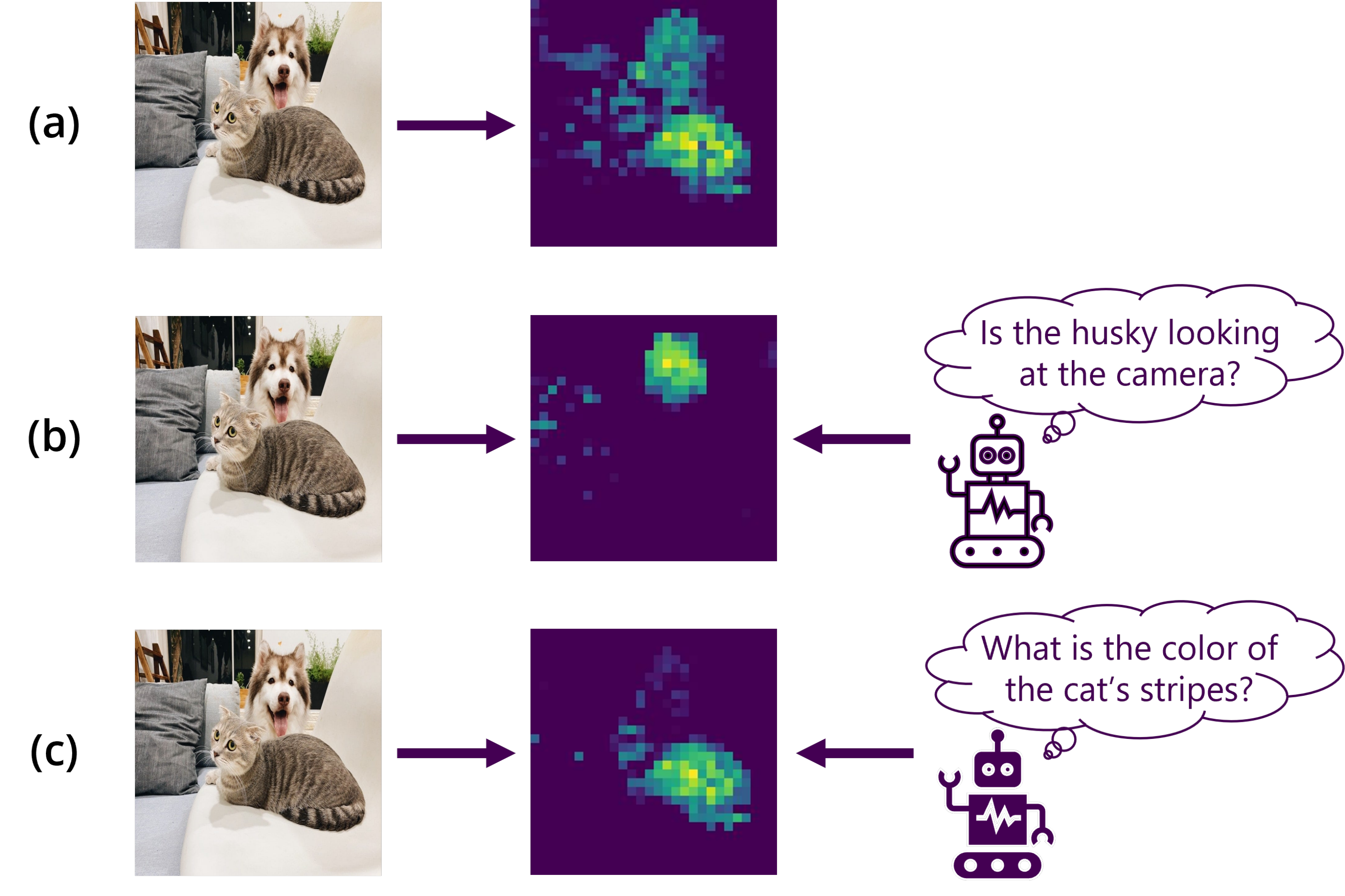}} \vspace{-2mm}
\caption{\textbf{Top-down \vs bottom-up attention}. (a) Bottom-up attention is stimulus-driven, \ie, any salient objects (dog and cat) in the image may attract attention. (b-c) Top-down attention is task-guided. For example, when the task is to answer a question about a specific object, the attention will only center on that object and ignore the others. In this way, a more focused representation can be extracted for the current goal.}
\label{fig:intro}
\end{center}
\vskip -0.4in
\end{figure}

Human visual attention is often \textit{task-guided}, \ie, we tend to focus on different objects when processing different tasks~\cite{zhaoping2014understanding,carrasco2011visual}. For example, when we answer different questions about one image, we only attend to the objects that are relevant to the question (\cref{fig:intro} (b-c)). This stands in contrast with the widely-used self-attention~\cite{dosovitskiy2020image}, which is completely \textit{stimulus-driven}, \ie, it highlights all the salient objects in the image without task-guided selection (\cref{fig:intro} (a)). While the stimulus-driven bottom-up attention has shown promising results in visual representation learning~\cite{caron2021emerging}, current vision transformers still lack the ability of task-guided top-down attention, which provides task-adaptive representation and potentially improves task-specific performances~\cite{anderson2018bottom,xu2015show,xu2016ask}.
Although some algorithms of top-down attention are proposed in the literature~\cite{pang2021tdaf,chen2021look,anderson2018bottom,xu2015show,xu2016ask}, they are incompatible with self-attention-based transformers and principled and unified designs are still missing.

Previous work~\cite{lee2002top,chikkerur2010and,borji2012object,rao2005bayesian,lee2003hierarchical} has studied the mechanism of top-down attention in human vision systems, hypothesizing top-down attention is a result of the human visual system performing Analysis by Synthesis (AbS). AbS~\cite{knill1996perception,yuille2006vision} is a classic idea that suggests the human visual perception depends on both the input image and a high-level prior about the latent cause of the image, and different priors can lead to different ways to perceive the same image (\eg, visual illusion~\cite{lee2003analysis} and bistable perception~\cite{schrater2006theory}). This is formulated as Bayesian inference $\max_\bfz p(\bfh | \bfz) p(\bfz)$, where $\bfh$ is the input image, and $\bfz$ is the latent representation. It is hypothesized that the high-level goal can be formulated as a prior to direct the low-level recognition of different objects through AbS, achieving top-down attention. Still, existing works~\cite{yu2004inference,chikkerur2010and,mirza2019introducing} are  conceptual and hardly guide model designs in practice.

In this work, we present a novel perspective on how AbS entails top-down attention, followed by a new Analysis-by-Synthesis Vision Transformer (\model) based on the findings.  We start from previous work~\cite{shi2022visual}, which shows that visual attention (\eg, self-attention) is functionally equivalent to sparse reconstruction which reconstructs the input using a dictionary containing templates of separate objects in the input.  
We show that AbS optimizes a similar \emph{sparse reconstruction} objective modulated by a top-down signal. The top-down signal depends on the prior and acts as a preference on which object templates to choose to reconstruct the input. Therefore, only the objects consistent with the high-level prior are selected, equivalent to top-down attention.

Inspired by the connection, we propose \model, a ViT~\cite{dosovitskiy2020image} model with prior-conditioned top-down modulation trained to approximate AbS in a variational way. \model contains a feedforward (encoding) and a feedback (decoding) pathway. The feedforward path is a regular ViT, and the feedback path contains linear decoders for each layer. Each inference starts with an initial feedforward run. The output tokens are manipulated by the prior and fed back through the decoders to each self-attention module as top-down input for the final feedforward pass (\cref{fig:model}). 

When only pretrained on ImageNet~\cite{deng2009imagenet}, which contains mostly single-object images, \model can attend to different objects in multi-object scenes controllably. For real-world applications, we observe consistent improvements from \model on Vision-Language tasks such as VQA~\cite{antol2015vqa} and zero-shot image retrieval, where language is used as a prior to guide attention. For tasks without a strong prior, such as ImageNet classification and semantic segmentation, \model can also serve as a general backbone and achieve substantial improvements. 
Additionally, the object-centric representation resulting from the top-down attention design enables 
better generalization to corrupted, adversarial, and out-of-distribution images. We hope this work can encourage future exploration of task-guided attention designs and visual representation learning. 

\section{Related Work}
\label{sec:relwork}

\minisection{Top-down visual attention}
endows us with the crucial ability to selectively collect information related to the behavioral goal. Several attempts have been made towards understanding the mechanism of top-down attention from experimental observations such as multiplicative tuning~\cite{mcadams1999effects} and contrast responses~\cite{martinez2002attentional,reynolds2009normalization} in V4, and extra-classical receptive fields in V1~\cite{angelucci2002circuits,sceniak2001visual,cavanaugh2002nature}. Other work tries to build a principled computational model for top-down attention~\cite{yu2004inference,chikkerur2010and,mirza2019introducing}.

Top-down attention has also found numerous applications in computer vision tasks where additional guidance (\eg, language) is available aside from the image. Previous work employs top-down attention for object detection~\cite{oliva2003top}, image captioning~\cite{xu2015show}, and visual question answering~\cite{xu2016ask,anderson2018bottom}. However, these algorithms are either incompatible with current self-attention-based models or show inferior performance, as indicated by our experiments. Other work ~\cite{lu2017knowing,lu2016hierarchical,yang2016stacked,dosovitskiy2020image} uses a feedforward model that takes both image and the high-level guidance (\eg, text tokens or \texttt{[cls]} token) as input, which we show is suboptimal compared to our top-down model design. 
Dou \etal~\cite{dou2022empirical}  propose to extract image and text features with separate encoders and combine them with a multi-modal fusion module during vision-language pretraining, which works better than using a single multi-modal feedforward model on vision language tasks. However, in this way, the visual encoder is still bottom-up. We show that augmenting it with the proposed top-down attention further improves model performance on standard benchmarks.

\minisection{Top-down attention explained as Analysis by Synthesis}.
Analysis by Synthesis (AbS) is hypothesized as a potential computational model behind top-down attention. \citet{lee2002top} starts from a Bayesian inference perspective and explains the top-down modulation in examples such as illusionary contours and shapes from shading. \citet{yu2004inference} focus on the top-down attention in Ponser's task~\cite{posner1980orienting} and build a hierarchical model where each layer corresponds to a computational step of Bayesian inference. Subsequent work~\cite{rao2005bayesian,chikkerur2010and} assumes each object is generated by an appearance variable and a location variable and uses Bayesian inference to perform spatial attention and feature attention. \citet{borji2012object} adopt a Dynamic Bayesian Network to simulate eye fixation in top-down attention. However, these models do not apply to practical designs in modern deep learning.

\minisection{Generative model for discriminative learning}.
It has been widely explored in using generative models to assist discriminative learning. Specifically, the belief that representation with strong generative capability can better capture the structure of visual signals has inspired numerous unsupervised learning algorithms, from the early Restricted Boltzmann Machine~\cite{hinton2006fast,hinton2006reducing} and Helmholtz Machine~\cite{dayan1995helmholtz}, to the following auto-encoder models such as DAE~\cite{vincent2010stacked} and VAE~\cite{kingma2013auto}. Recent work~\cite{he2022masked,tong2022unsupervised} has shown impressive results on generative unsupervised learning. Generative models can also help with supervised learning, \eg, by refining object detection~\cite{lin2017feature} or detecting errors in semantic segmentation~\cite{xia2020synthesize}. Feedforward models with generative feedback are also more robust to input corruptions~\cite{huang2020neural}. In our work, \model also contains a generative feedback path that is able to refine the intermediate representation and attention and thus improves the performance.

\section{Preliminaries: Attention as Sparse Reconstruction}
\label{sec:att_emerge}

\citet{shi2022visual} show that a sparse reconstruction (SR) module functionally resembles visual attention. An SR module takes an input $\bfx \in \bbR^d$ %
and outputs $\bfz = \bfP\tbfu^\ast$ where $\bfP \in \bbR^{d \times d^\prime}$ is the dictionary and $\tbfu^\ast$ is the sparse code, \ie, 
\begin{equation}\small
\label{eq:sr}
    \tbfu^\ast = \argmin_{\tbfu \in \bbR^{d^\prime}}  \frac{1}{2} ||\bfP \tbfu - \bfx||^2_2 + \lambda ||\tbfu||_1.
\end{equation}
Each atom (column) of $\bfP$ contains a template pattern and each element in $\tbfu$ is the activation of the corresponding template. The objective is to reconstruct the input using as few templates as possible. To solve \cref{eq:sr}, one may adopt a first-order optimization \cite{rozell2008sparse,shi2022visual} with dynamics at time $t$ of
\begin{equation}\small
\label{eq:sr_dyn}
    \dv{\bfu}{t} \propto -\bfu - (\bfP^T \bfP - \mathbf{I}) \tbfu  + \bfP^T \bfx,
\end{equation}
where the optimization is over an auxiliary variable $\bfu$ and $\tbfu = \mathit{g}_\lambda (\bfu) = \mathit{sgn}(\bfu) (|\bfu| - \lambda)_+$ with $\mathit{sgn}(\cdot)$ as the sign function and $(\cdot)_+$ as ReLU %
Here $\bfu$ is activated by the template matching $\bfP^T \bfx$ between the dictionary and the input, and different elements in $\bfu$ inhibit each other through $- (\bfP^T \bfP - \mathbf{I}) \tbfu$ to promote sparsity.

To see the connection between visual attention and sparse reconstruction, recall that attention in the human visual system is achieved via two steps~\cite{desimone1995neural}: (i) \textit{grouping} features into separate objects or regions, and (ii) \textit{selecting} the most salient objects or regions while repressing the distracting ones. A similar process is also happening in SR, \ie, if each atom in $\bfP$ is a template of every single object, then each element in $\bfu$ groups the input features belonging to that object through $\bfP^T \bfx$, while the sparsity constraint promoted by the lateral inhibition $- (\bfP^T \bfP - \mathbf{I}) \tbfu$ selects the object that is most activated. As shown in~\cite{shi2022visual}, SR modules achieve similar attention effects as self-attention (SA)~\cite{vaswani2017attention} while being more robust against image corruptions.

Interestingly, it is also pointed out in~\cite{shi2022visual} that under certain constraints (\eg, the key and query transform is the same), SA can be viewed as solving a similar SR problem but without sparsity. After adding the sparsity back, SA is an approximation of 

\vspace{-0.5em}
\begin{footnotesize}
\begin{empheq}[left=\empheqlbrace]{align}
\label{eq:sa_sr}
    \tbfU^\ast &= \argmin_{\tbfU} \frac{1}{2} ||\Phi(\bfK)\tbfU - \bfV||_2^2 + \lambda||\tbfU||_1, \\
    \bfZ &= \Phi(\bfQ)\tbfU^\ast,
\end{empheq}
\end{footnotesize}\noindent
where $\bfQ, \bfK, \bfV \in \bbR^{(hw)\times c}$ are the query, key, and value matrices, $\Phi(\bfQ), \Phi(\bfK) \in \bbR^{(hw) \times d^\prime}$ are the random features~\cite{choromanski2020rethinking} that approximate the softmax kernel $\Phi(\bfQ)_i \Phi(\bfK)_j^T \approx e^{\bfQ_i \bfK_j^T}$, $\tbfU^\ast \in \bbR^{d^\prime \times c}$ is the sparse code and $\bfZ$ is the output. This provides a novel perspective on the mechanism of SA, \ie, it is solving a channel-wise sparse reconstruction of the value matrix $\bfV$ using an input-dependent dictionary $\Phi(\bfK)$. Visualization of $\Phi(\bfK)$ shows each atom contains a mask for one single object or region, which means that SA is trying to reconstruct the input with as few masks as possible, thus only the salient objects are selected and highlighted (\cref{fig:dict} (a)).
\section{Top-Down Attention from AbS}
\label{sec:deriv}

We consider top-down visual attention from an Analysis by Synthesis (AbS) view of vision. We start from the hierarchical AbS formulation of visual perception (\cref{sec:abs}) and show that it is equivalently optimizing a sparse reconstruction objective that is modulated by a top-down signal, thus entailing top-down attention (\cref{sec:td_from_abs}).

\subsection{Hierarchical AbS}
\label{sec:abs}

AbS formulates visual perception as a Bayesian inference process. Given the image generation process $p(\bfh | \bfz)$ and a prior $p(\bfz)$, where $\bfh$ is the image and $\bfz$ is the latent code, AbS finds $\bfz^\ast = \argmax_{\bfz} p(\bfh | \bfz) p(\bfz)$.

In this work, we assume the generation is hierarchical, \ie, $\bfz_L \to \bfz_{L-1} \to  \cdots \to \bfz_1 \to \bfh$, where $\bfz_\ell$ is the latent at $\ell$-th layer. The MAP estimation is
\begin{equation}\small
\label{eq:abs}
    \bfz_L^\ast, \cdots, \bfz_1^\ast = \argmax_{\bfz_L, \cdots, \bfz_1} p(\bfh | \bfz_1) \cdots p(\bfz_{L-1} | \bfz_L) p(\bfz_L).
\end{equation}

For each generation process $\bfz_{\ell+1} \to \bfz_\ell$ between layer $\ell$ and $\ell+1$, we further assume that $\bfz_\ell$ is constructed by a sparse code $\tbfu_\ell$ which is generated from $\bfz_{\ell+1}$ via a non-linear function $\mathit{g}_\ell(\cdot)$, \ie,

\vspace{-0.8em}
\begin{footnotesize}
\begin{align}
    \tbfu_\ell &\sim p(\tbfu_\ell | \bfz_{\ell+1}) \propto \exp\{-\frac{1}{2}||\bfP_\ell \tbfu_\ell - \mathit{g}_\ell(\bfz_{\ell+1})||_2^2 - \lambda||\tbfu_\ell||_1\} \label{eq:gen_1}\\
    \bfz_\ell &= \bfP_\ell \tbfu_\ell \label{eq:gen_2},
\end{align}
\end{footnotesize}\noindent
where $\bfP_\ell$ is the dictionary. Intuitively, it first generates $\mathit{g}_\ell(\bfz_{\ell+1})$ as a blurry and noisy version of $\bfz_\ell$, then find the sparse code $\tbfu_\ell$ to construct a sharper and cleaner version.

Since $\bfz_\ell$ is decided by $\tbfu_\ell$, it suffices to optimize the MAP estimation over $\{\tbfu_\ell\}_{\ell=1}^L$, \ie,
\begin{equation}\small
\label{eq:abs_2}
    \tbfu_L^\ast, \cdots, \tbfu_1^\ast = \argmax_{\tbfu_L, \cdots, \tbfu_1} p(\bfh | \tbfu_1) \cdots p(\tbfu_{L-1} | \tbfu_L) p(\tbfu_L).
\end{equation}
Solving \cref{eq:abs_2} by simple gradient ascent (of the logarithm) gives the dynamics
\begin{equation}\small
\label{eq:abs_dyn}
    \dv{\tbfu_\ell}{t} \propto \nabla_{\tbfu_\ell} \log p(\tbfu_{\ell-1} | \tbfu_\ell) + \nabla_{\tbfu_\ell} \log p(\tbfu_\ell | \tbfu_{\ell+1})
\end{equation} 
where $\tbfu_\ell$ is affected by both $\tbfu_{\ell-1}$ and $\tbfu_{\ell+1}$.

\subsection{Top-Down Attention from AbS}
\label{sec:td_from_abs}

\begin{figure}[t]
\begin{center}
\centerline{\includegraphics[width=0.95\columnwidth]{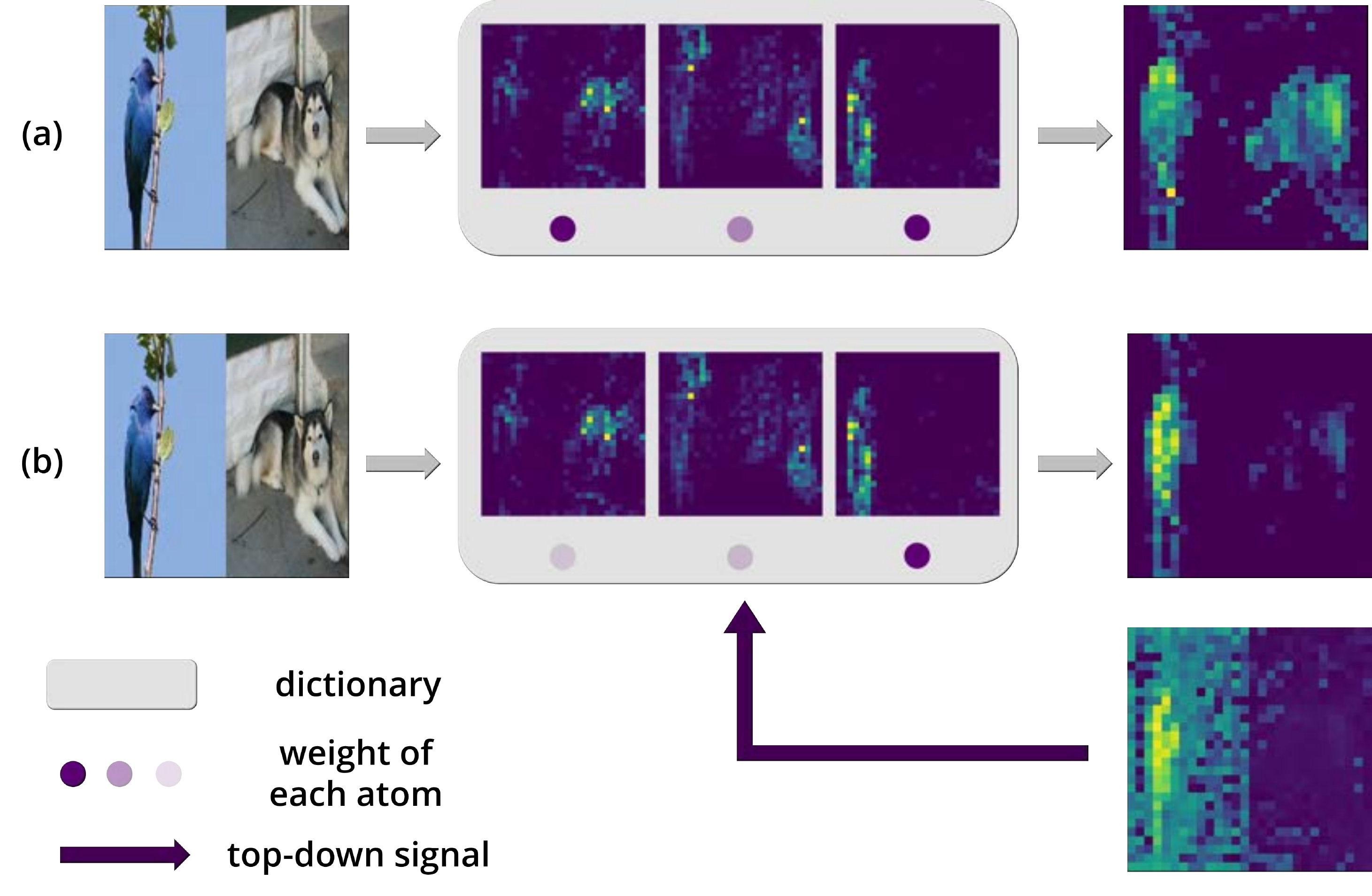}}
\caption{(a) Each atom in the dictionary contains masks for separate objects or regions. The sparse reconstruction tries to use as few masks as possible to reconstruct the input feature map, thus only the salient objects are highlighted. (b) The top-down signal $\bfx^{td}_\ell$ puts a bias on the weights of the atoms so that only the objects that agree with $\bfx^{td}_\ell$ are selected.}
\label{fig:dict}
\end{center}
\vskip -0.3in
\end{figure}

From AbS (Eq. (\ref{eq:gen_1}-\ref{eq:abs_dyn})) we can derive the dynamics of $\tbfu_\ell$ as 
\begin{equation}\footnotesize
\label{eq:abs_dyn2}
    \dv{\tbfu_\ell}{t} \propto \nabla_{\tbfu_\ell} \left( -\frac{1}{2} ||\bfP_\ell \tbfu_\ell - (\bfx_\ell^{bu} + \bfx_\ell^{td}) ||_2^2  - \lambda ||\tbfu_\ell||_1 - \mathit{r}_\ell(\tbfu_\ell) \right)
\end{equation}
where $\bfx_\ell^{td} = \mathit{g}_\ell (\bfz_{\ell+1})$ is the top-down signal and $\bfx_\ell^{bu} = \mathit{f}_\ell (\bfz_{\ell-1}) = \bfJ_{\mathit{g}_{\ell-1}}^T \bfz_{\ell-1} $ is the bottom-up signal where $\bfJ_{\mathit{g}_{\ell-1}}$ is the jacobian of $\mathit{g}_{\ell-1}(\bfP_\ell \tbfu_\ell)$, and $\mathit{r}_\ell(\tbfu_\ell) = ||\mathit{g}_{\ell-1}(\bfP_\ell \tbfu_\ell)||_2^2$ is an additional regularization. Details of the derivation are pushed back to Appendix. One may notice from \cref{eq:abs_dyn2} that, in AbS  each layer is solving a similar sparse reconstruction problem as in \cref{eq:sr} but with the input of $\bfx_\ell^{bu} + \bfx_\ell^{td}$, thus simulating attention that is modulated by both bottom-up and top-down signals. This can also be observed by turning \cref{eq:abs_dyn2} into
\begin{equation}\small
    \dv{\tbfu_\ell}{t} \propto -\bfu_\ell  - (\bfP_\ell^T \bfP_\ell - \mathbf{I})\tbfu_\ell + \bfP_\ell^T \bfx_\ell^{bu} + \bfP_\ell^T \bfx_\ell^{td} - \nabla \mathit{r}_\ell(\tbfu_\ell).
\end{equation}
Comparing with \cref{eq:sr_dyn}, here $\tbfu_\ell$ is steered by an additional term $\bfP_\ell^T \bfx_\ell^{td}$ that acts as a bias on which atom in $\bfP_\ell$ to choose. For example, if atoms in $\bfP_\ell$ are templates of separate objects (like in self-attention), then $\bfP_\ell^T \bfx_\ell^{td}$ highlights the objects that are consistent with the top-down signal (\cref{fig:dict} (b)).

This implies an AbS system naturally entails top-down attention. Intuitively, the prior reflects which objects the output $\bfz_L$ should highlight. Then the affected $\bfz_L$ is fed back to layer $L-1$ through $\mathit{g}_{L-1}$, as a top-down signal to direct which objects to select in layer $L-1$. The same process repeats until the first layer. Different priors will direct the intermediate layers to select different objects, achieving top-down attention. 

Interestingly, if we consider the analogy between self-attention and sparse reconstruction, \cref{eq:abs_dyn2} leads to a smooth way of building a top-down version of self-attention, \ie, we only need to add a top-down signal to the value $\bfV$, while keeping other parts such as $\bfQ$ and $\bfK$ (which decides the dictionary) untouched. We will make it clearer in \cref{sec:implem}.

\section{Analysis-by-Synthesis Vision Transformer}
\label{sec:implem}

\begin{figure*}[t]
\begin{center}
\centerline{\includegraphics[width=2\columnwidth]{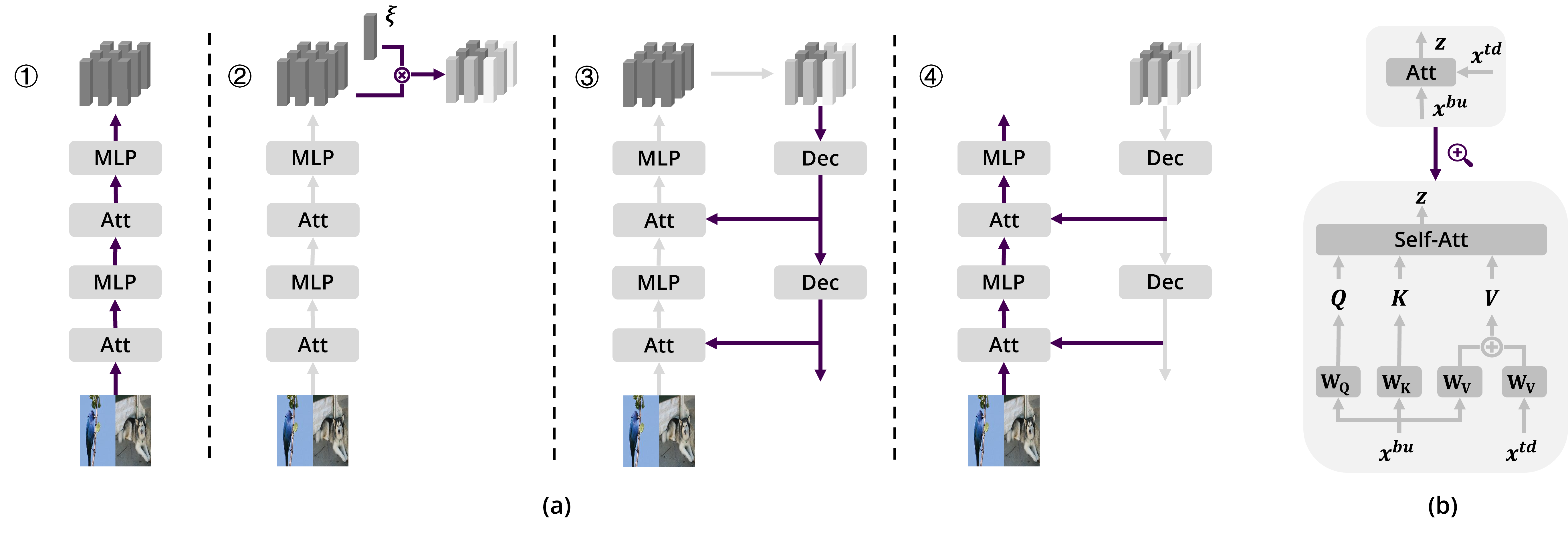}}
\caption{\textbf{Design of \model}. (a) Four steps to every single inference. The operations in each step are colored as purple and others as gray. \model first passes the image through the feedforward path. The output tokens are then reweighted by their similarity with the prior vector $\xi$ and fed back through the decoders to each self-attention module as the top-down input for the final feedforward run. (b) The top-down input to self-attention is added to the value matrix while other parts stay the same.   }
\label{fig:model}
\end{center}
\vskip -0.3in
\end{figure*}

Inspired by the connection between top-down attention and AbS, we propose to achieve top-down attention by building a vision transformer that performs AbS (\cref{eq:abs}), \ie, if the network has input $\bfh$ and latent representation $\bfz_\ell$ after each layer $\ell$ (which means $\bfz_L$ is the output), the final latent representation should approximate $\bfz_1^\ast, \cdots, \bfz_L^\ast$. Since directly solving \cref{eq:abs} requires an iterative optimization which would be extremely costly, in this work, we adopt a variational approximation to \cref{eq:abs}. Specifically, we optimize a variational loss
\begin{equation}\footnotesize
\label{eq:var_loss}
\begin{split}
    \mathcal{L}_{var}&= -\sum_{\ell=0}^{L-1} \log p(\bfz_\ell | \bfz_{\ell+1}) - \log p(\bfz_L) \\
    &= \sum_{\ell=0}^{L-1} \left(\frac{1}{2}||\bfP_\ell \tbfu_\ell - \mathit{g}_\ell(\bfz_{\ell+1})||_2^2 + \lambda||\tbfu_\ell||_1 \right) - \log p(\bfz_L) 
\end{split}
\end{equation}
where $\bfz_0 = \bfh$. However, as stated below, there are several caveats we need to work around when training a network with \cref{eq:var_loss} in real-world tasks.

\minisection{The sparsity regularization}. Since the practical model we build in this work is based on self-attention (\cref{sec:design}), which neither has a sparsity constraint nor solves the SR explicitly~\cite{shi2022visual}, we remove the sparsity regularization by setting $\lambda = 0$, which makes $-\log p(\bfz_\ell | \bfz_{\ell+1}) = \frac{1}{2}||\bfP_\ell \tbfu_\ell - \mathit{g}_\ell(\bfz_{\ell+1})||_2^2 = \frac{1}{2}||\bfz_\ell - \mathit{g}_\ell(\bfz_{\ell+1})||_2^2$.

\minisection{Jointly training the decoder $\mathit{g}_\ell$}. Normally, optimizing \cref{eq:var_loss} requires knowing the generation process $\mathit{g}_\ell$ beforehand, which in our case is unknown. This can be addressed by training $\mathit{g}_\ell$ jointly with the whole network, similar to VAE~\cite{kingma2013auto}. It is natural to use $\mathit{g}_\ell$ also as the feedback path of the network, as shown in \cref{sec:design}.

\minisection{Trade-off between the generative and discriminative power}. The variational loss forces each $\bfz_{\ell+1}$ to be capable of generating $\bfz_\ell$. However, we find empirically that enforcing a strong generative power on the feature will harm its discriminative power in the setting of supervised learning. %
To address this, for each term $-\log p(\bfz_\ell| \bfz_{\ell+1})$ we stop the gradient on $\bfz_\ell$ and $\bfz_{\ell+1}$, \ie, $-\log p(\bfz_\ell| \bfz_{\ell+1}) = \frac{1}{2}||\mathit{sg}(\bfz_\ell) - \mathit{g}_\ell(\mathit{sg}(\bfz_{\ell+1}))||_2^2$, where $\mathit{sg}(\cdot)$ is stop-gradient. In this way, only the decoder $\mathit{g_\ell}$ receives the gradient.

\minisection{The variable prior}. Rigorously speaking, variational methods only approximate AbS with a fixed prior $p(\bfz_L)$. However, top-down attention should be able to flexibly attend to different objects by changing different priors. The question is, how can we learn a variational model that generalizes to different priors? In this work, we adopt a simple trick called Meta-amortized VI~\cite{wu2020meta}. Concretely, we assume the prior $p_\xi(\bfz_L)$ depends on some parameter $\xi$, which can be a sentence or a class prototype cueing what objects to look at in the image. Then we make the model adaptable to $\xi$ during inference to approximate AbS with prior $p_\xi(\bfz_L)$ given any $\xi$. See the design details in \cref{sec:design}.

\vspace{1.3mm}
After applying these tricks, our variational loss becomes 
\begin{equation}\small
\label{eq:var_loss_2}
    \mathcal{L}_{var} = \frac{1}{2} \sum_{\ell=0}^{L-1} ||\mathit{sg}(\bfz_\ell) - \mathit{g}_\ell(\mathit{sg}(\bfz_{\ell+1}))||_2^2  - \log p_\xi(\bfz_L),
\end{equation}
which contains layer-wise reconstruction loss and a prior loss. We also try cosine similarity instead of $\ell_2$ distance for reconstruction and get similar results. In \cref{sec:design}, we will show how to build a ViT with prior-conditioned top-down modulation and train it with \cref{eq:var_loss_2}.

\subsection{\model Design}
\label{sec:design}

\cref{fig:model} (a) shows the proposed \model which is built upon ViT~\cite{dosovitskiy2020image}. Every single inference consists of 4 steps: (i) pass the image through the feedforward encoder, (ii) modulate the output tokens with a prior vector $\xi$, (iii) send the tokens back through the feedback decoder to intermediate layers, and (iv) run the feedforward path again but with each self-attention layer also receiving the top-down tokens as input.

Within the whole pipeline, the feedforward encoder has the same architecture as regular ViT. For the feedback path, we use a single token-wise linear transform for each layer-wise decoder $\mathit{g}_\ell$. The design of token modulation with prior $\xi$ and the self-attention with top-down input are introduced below:

\minisection{Design of token modulation with $\xi$}. The purpose is to modify the tokens to carry the information about the prior $p_\xi$ when fed back to the network. The prior is parameterized by $\xi$, which may be a language embedding or a class prototype telling the network which objects to look at. Therefore, we instantiate the modulation as a simple spatial reweighting, \ie, $\bfz_L^i \to \alpha \cdot \mathit{sim}(\xi, \bfz_L^i) \cdot \bfz_L^i$, where $\bfz_L^i$ is the $i$-th output token, $\mathit{sim}$ is the cosine similarity clamped to $[0, 1]$, and $\alpha$ is a scaling factor controlling the scale of the top-down signal, which is set to 1 by default. In this way, only the tokens with high  similarity to $\xi$ are sent back, and others are (softly) masked out. Note that the design here is for simplicity and may not be suitable for general usage. For example, when dealing with transparent images where two objects overlap, spatial reweighting cannot separate two objects away.
    
\minisection{Design of self-attention with top-down input}. From the analogy between self-attention and sparse reconstruction (\cref{eq:sa_sr}), the value matrix in SA corresponds to the reconstructed input signal, and the query and key serve as the dictionary. Since the top-down attention in AbS (\cref{eq:abs_dyn2}) adds a top-down signal to the input while keeping the dictionary untouched, it is natural to design the top-down version of self-attention by simply adding the top-down signal to the value and keep query and key as the same, as illustrated in \cref{fig:model} (b). We will show in \cref{sec:exp_model_design} that this is better than an arbitrary design where we add the top-down signal to the query, key, and value.

\vspace{1.3mm}
In this paper, we focus on supervised learning and train the model on two types of tasks. One is Vision-Language (V\&L) tasks such as VQA and zero-shot image retrieval, where the language acts as a prior to cue the model where to look at. The other one is image understanding, such as ImageNet classification and semantic segmentation, which do not have a specific prior. When training the network, we optimize the supervised loss as well as the variational loss (\cref{eq:var_loss_2}), \ie, 
\begin{equation}\small
    \mathcal{L} =  \frac{1}{2} \sum_{\ell=1}^L ||\mathit{sg}(\bfz_\ell) - \mathit{g}_\ell(\mathit{sg}(\bfz_{\ell+1}))||_2^2  - \log p_\xi(\bfz_L) + \mathcal{L}_{sup},
\end{equation}
where $\bfz_\ell$ is the $\ell$-th layer's output after the whole inference cycle, $\mathit{sg}$ is stop-gradient, and $\mathit{g}_\ell$ is the $\ell$-th layer's decoder. The form of prior $p_\xi$ depends on the task. For V\&L tasks, $\xi$ is the text embedding and we use a CLIP-style prior~\cite{radford2021learning}:
\begin{equation}\small
\label{eq:clip_loss}
    p_\xi(\bfz_L) = \frac{\exp\{\xi^T \bfz_L\}}{\exp\{\xi^T \bfz_L\} + \sum_k \exp\{\xi^T \bfz_-^k\}},
\end{equation}
where the negative samples $\bfz_-^k$ are the output from other images. For image classification and segmentation where no specific prior is available,  we set $\xi$ as a trainable query vector that is independent of the input image, and we choose an uninformative prior that does not contribute to the gradient, \ie, $\nabla \log p_\xi(\bfz_L) = 0$.

\begin{figure}[t]
\begin{center}
\centerline{\includegraphics[width=0.85\columnwidth]{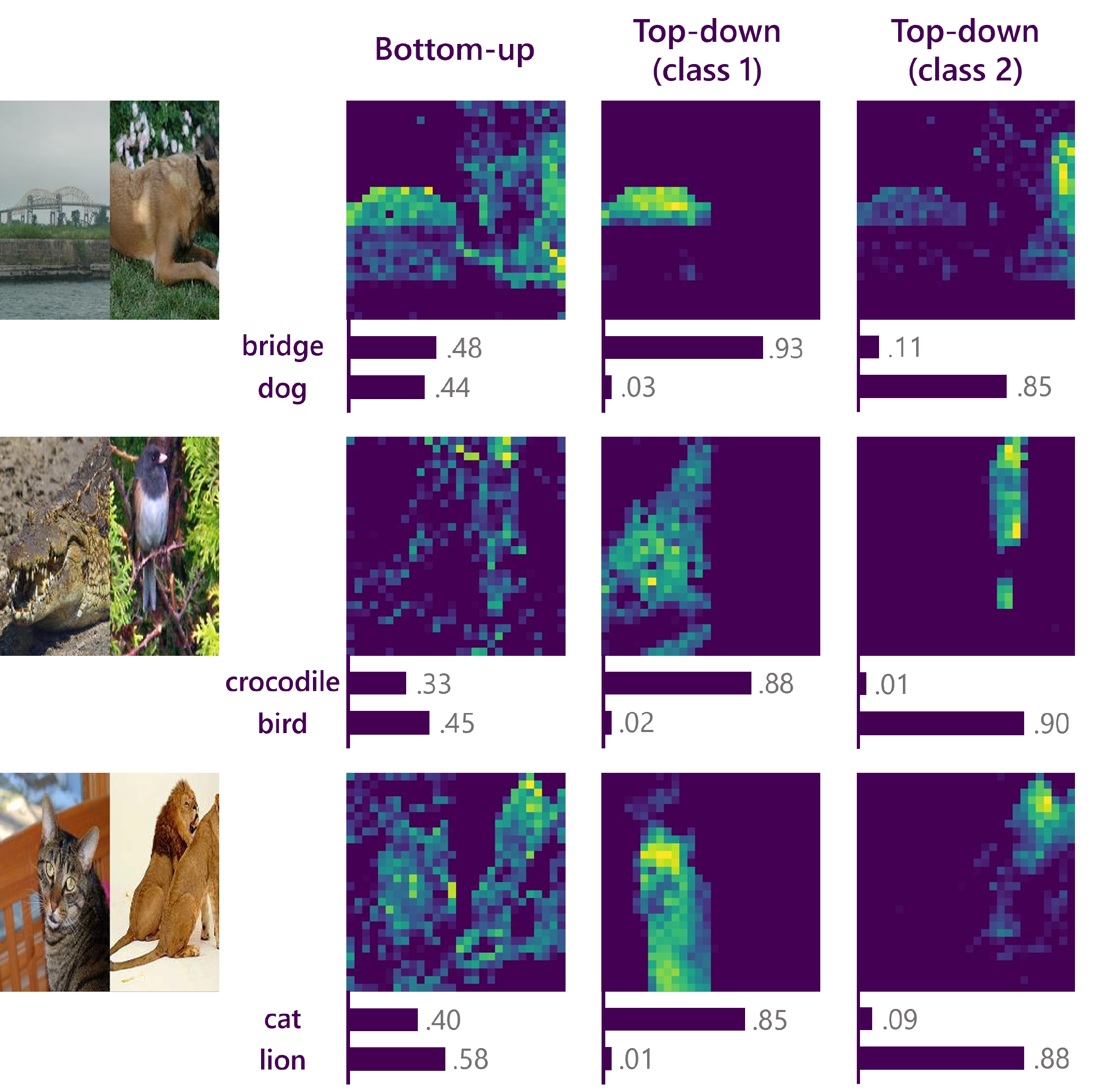}}
\caption{Controllable top-down attention in multi-object images. For each image, bottom-up attention will highlight both objects. In contrast, we can use different class prototypes as the prior to control the top-down attention to focus on different objects, and the classification result also changes accordingly.}
\label{fig:spatial_bistable}
\end{center}
\vskip -0.3in
\end{figure}

\section{Experiments}
\label{sec:exp}

In this section, we first show that \model achieves controllable top-down attention in multi-object scenes (\cref{sec:exp_td_att}). Then we test \model on Vision-Language tasks such as VQA and zero-shot image retrieval (\cref{sec:exp_vl}), and also on ImageNet classification and model robustness (\cref{sec:exp_imagenet}), as well as semantic segmentation (\cref{sec:exp_segmentation}). Finally, we analyze specific designs of \model in \cref{sec:exp_model_design}.

\minisection{Datasets}. For VQA, we use VQAv2~\cite{goyal2017making} for training and testing and compare the attention map with human attention collected by VQA-HAT~\cite{das2017human}. For zero-shot image retrieval, we use Flickr30K~\cite{plummer2015flickr30k}. For image classification, we train and test on ImageNet-1K (IN)~\cite{deng2009imagenet}, and also test on corrupted images from IN-C~\cite{hendrycks2019benchmarking}, adversarial images from IN-A~\cite{hendrycks2021natural}, and out-of-distribution images from IN-R~\cite{hendrycks2021many} and IN-SK~\cite{wang2019learning}. For semantic segmentation, we test on PASCAL VOC~\cite{pascal-voc-2012}, Cityscapes~\cite{Cordts2016Cityscapes}, and ADE20K~\cite{zhou2017scene}.

\minisection{Experimental setup}. We compare several baselines for goal-directed attention: (i) \textbf{PerceiverIO}~\cite{jaegle2021perceiver} uses $e_\xi(\cdot)$ to reweight the tokens from feedforward output just like in \model, but directly outputs the reweighted tokens without any feedback, (ii) \textbf{MaskAtt} uses the same soft mask for reweighting the output tokens to reweight the value tokens in intermediate self-attention modules, instead of adding the top-down tokens on them, (iii) \textbf{Feedback} directly feeds back the output tokens without reweighting. For V\&L tasks, we use the METER~\cite{dou2022empirical} framework, which contains a vision backbone, a language backbone, and a multimodal fusion module. We use ViT~\cite{dosovitskiy2020image} as the vision backbone and replace it with \model or the baseline models. For image classification, we try the backbones of ViT, RVT~\cite{mao2022towards}, and FAN~\cite{zhou2022understanding}, which is state of the art on ImageNet and robustness benchmarks. The scaling factor $\alpha$ is set as $1$ during ImageNet pretraining and evaluation and set as $10$ for finetuning on V\&L tasks because we find \model pretrained on supervised single-object classification only learns weak top-down attention in multi-object scenes (\cref{sec:lim_weak_td}). See the Appendix for additional implementation details.

\subsection{Controllable Top-Down Attention of \model}
\label{sec:exp_td_att}

\begin{figure}[t]
\begin{center}
\centerline{\includegraphics[width=0.85\columnwidth]{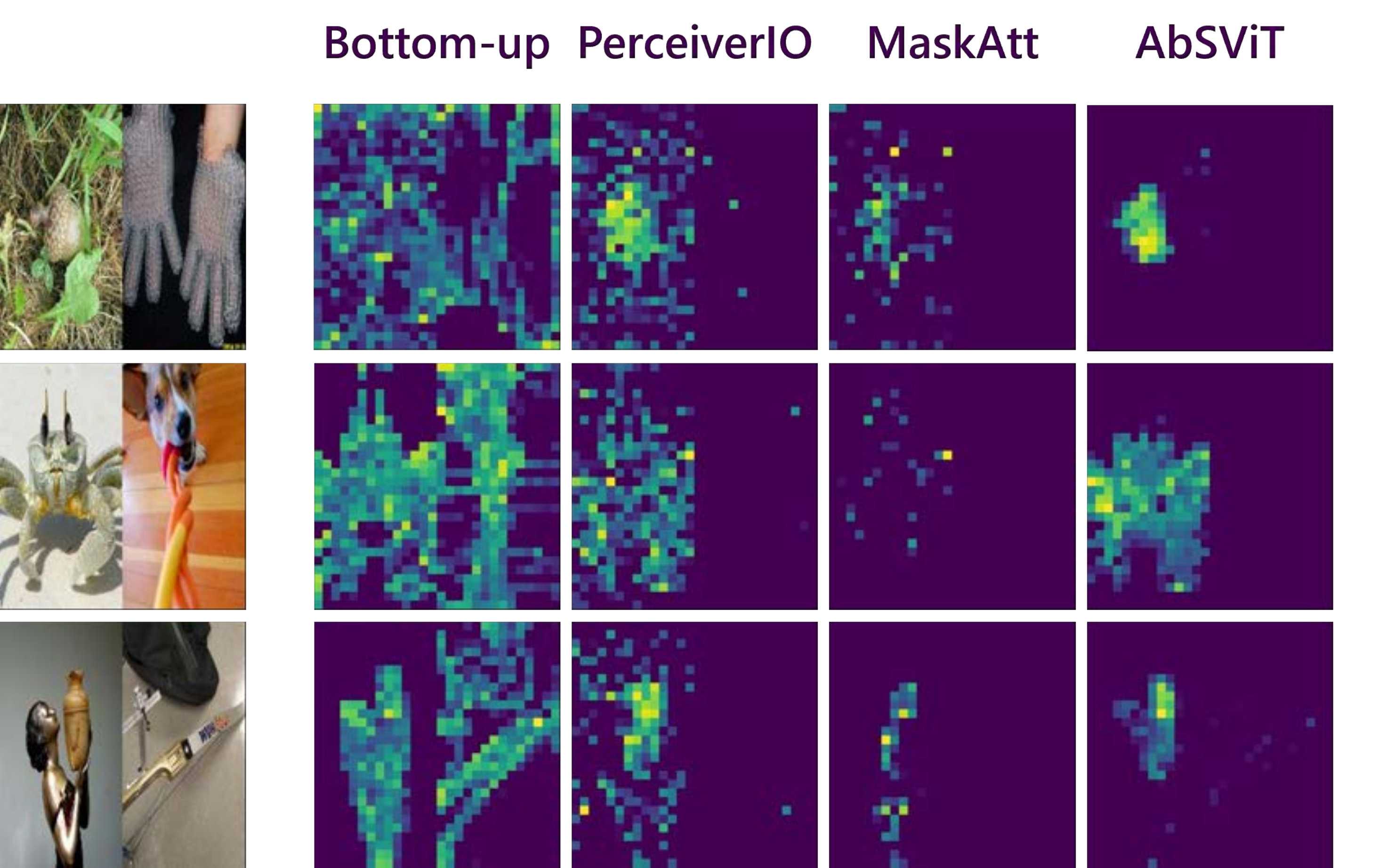}}
\caption{Comparison between different top-down attention algorithms. Prior corresponds to the left image. \model has cleaner attention map than other baselines.}
\label{fig:att_compare}
\end{center}
\vskip -0.3in
\end{figure}

To test the top-down attention in multi-object images, we take a \model pretrained on ImageNet (\cref{sec:exp_imagenet}) and create multi-object images by randomly sampling two images from ImageNet and concatenating them side by side. To control the top-down attention, we use the class prototype (from the last linear layer) of the two classes as $\xi$. Since in regular ViT, the class prototypes only align with the \texttt{[cls]} token but not with other output tokens, here we use a ViT with global average pooling. We set $\alpha=10$.

To compare the bottom-up and top-down attention, we visualize the norm of output tokens from ViT and \model for each class. As shown in \cref{fig:spatial_bistable}, bottom-up attention highlights both objects while only the target object is selected by top-down attention. Consequently, the classification result, which has a tie between two classes when no prior is available, is biased towards the target class when we turn on the prior. This indicates \model has the ability to control its attention on different objects given different priors. We also compare the top-down attention of \model with several baselines (\cref{fig:att_compare}). We can see that the attention of PerceiverIO focuses coarsely on the target object but is noisy, possibly because it lacks a feedback mechanism. MaskAtt, on the other hand, tends to miss parts of the object, implying that masking attention is less suitable for ViTs.

\subsection{\model for Vision-Language Tasks}
\label{sec:exp_vl}

\begin{table}[t]
  \centering
  \begin{footnotesize}
  \begin{tabular}{@{}lccccc@{}}
    {\multirow{2}{*}{Model}} & \multicolumn{2}{c}{VQAv2} & \multicolumn{3}{c}{Flickr-Zero-Shot} \\
     &  test-dev & test-std & IR@1 & IR@5 & IR@10 \\
    \midrule    
    BEiT-B-16~\cite{bao2021beit} & 68.45 & - & 32.24 & - & - \\
    CLIP-B-32~\cite{radford2021learning} & 69.69 & - & 49.86 & - & - \\
    \midrule
    ViT-B & 67.89 & 67.92 & 42.40 & 77.18 & 86.82 \\
    - PerceiverIO & 67.87 & 67.93 & 42.52 & 76.92 & 86.73 \\
    - Feedback & 67.99 & 68.13 & 42.04 & 77.38 & 86.90 \\
    - MaskAtt & 67.53 & 67.51 & 41.89 & 76.53 & 86.78 \\
    - \model & \textbf{68.72} & \textbf{68.78} & \textbf{45.28} & \textbf{77.98} & \textbf{87.52} \\
\bottomrule
  \end{tabular}
  \caption{Comparison of different top-down attention algorithms on VQA and zero-shot image retrieval. \model achieves consistent improvements on both tasks.}
  \label{tab:vl}
  \end{footnotesize}
\end{table}

\begin{figure}[t]
\begin{center}
\centerline{\includegraphics[width=0.9\columnwidth]{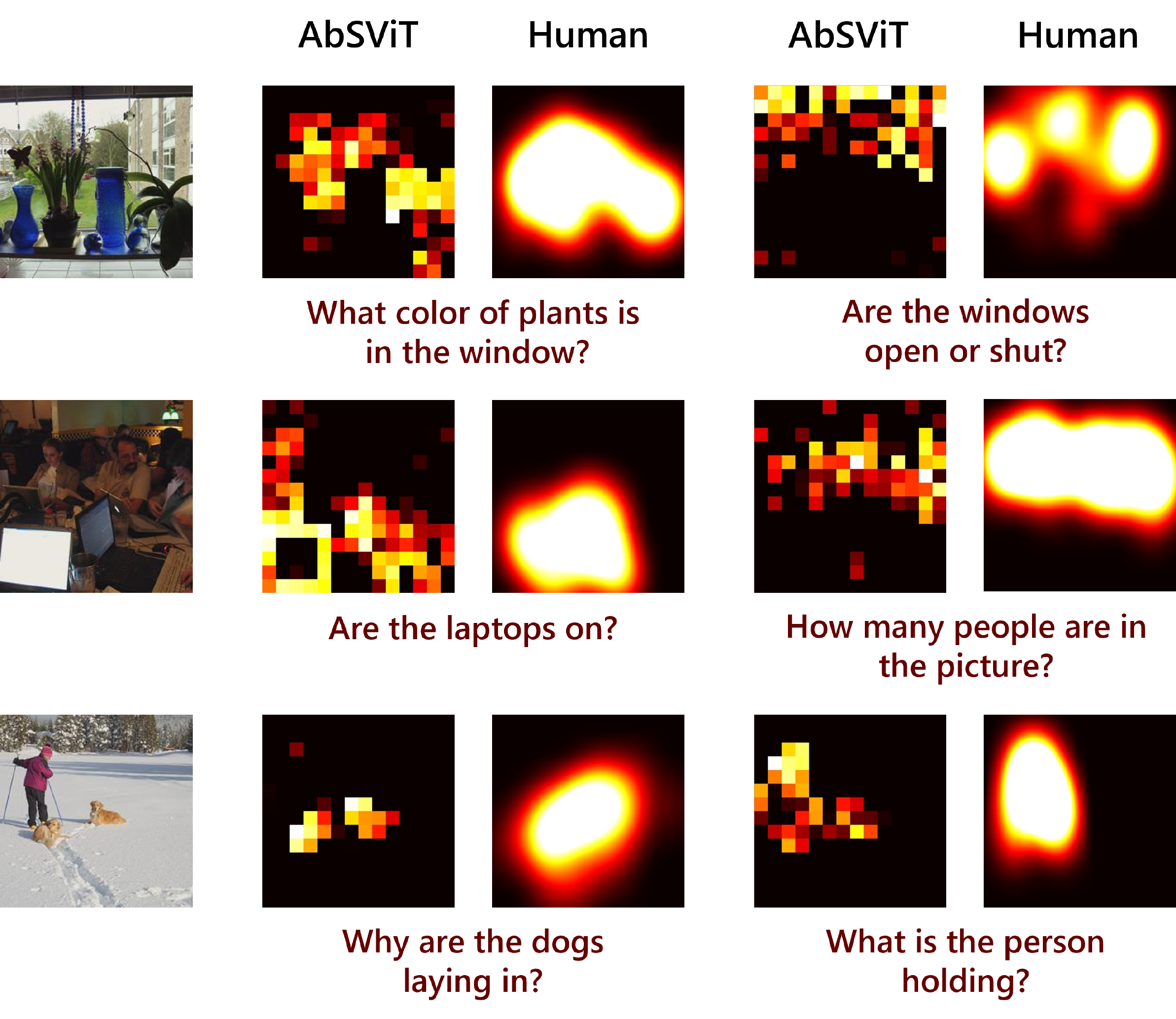}}
\caption{Comparison of attention map from \model and human attention on VQA. \model's attention is adjustable to different questions and is consistent with human attention. }
\label{fig:vqa_hat}
\end{center}
\vskip -0.3in
\end{figure}

We test AbSViT on two V\&L tasks, VQA, and zero-shot image retrieval. We use the METER framework and replace the vision backbone with ViT-B, AbSViT-B, and other baselines. All the vision backbones are pretrained on ImageNet (\cref{sec:exp_imagenet}). Results are shown in \cref{tab:vl}. 

On VQAv2, \model surpasses the baselines on both test splits and reaches the same performance as the unsupervised model (BEiT-B). At the same time, PerceiverIO has no improvement over ViT, probably because the multimodal fusion in METER can already perform token reweighting. The pure feedback network helps a little, mainly due to the feature refinement during the feedback loop. It is worth noticing that MaskAtt, a strategy frequently used in previous work, actually hurts performance when added to the vision transformer. On zero-shot image retrieval, \model also has higher performance than all other baselines. Especially, it has an improvement of $\sim3\%$ over bottom-up ViT on IR@1.

We also visualize the attention map of \model on VQA and compare it to human attention. As shown in \cref{fig:vqa_hat}, \model can adjust its attention to the objects related to the question. The attention map is also consistent with human attention.%
Nevertheless, the attention map of \model is still not precise enough. For example, in the last example, when the question is ``What is the person holding?'', the top-down attention highlights both the person and the dogs. Since the model is only pretrained on ImageNet, it may be further improved by CLIP~\cite{radford2021learning} pretraining.

\subsection{Image Classification and Robustness}
\label{sec:exp_imagenet}

\begin{table}[t]
  \centering
  \begin{scriptsize}
  \begin{tabular}{lcccccc}
    Model & P/F & Clean & IN-C ($\downarrow$) & IN-A & IN-SK & IN-R \\
    \midrule
    
    PiT-Ti~\cite{heo2021rethinking} & 5/0.7 & 72.9 & 69.1 & 6.2 & 34.6 & 21.6 \\
    ConViT-Ti~\cite{d2021convit} & 6/1.4 & 73.3 & 68.4 & 8.9 & 35.2 & 22.4 \\
    PVT-Ti~\cite{wang2021pyramid} & 13/1.9 & 75.0 & 79.6 & 7.9 & 33.9 & 21.5 \\
    GFNet-Ti~\cite{rao2021global} & 8/1.3 & 74.6 & 65.9 & 6.3 & 40.4 & 27.0 \\
    \midrule
    ViT-Ti~\cite{dosovitskiy2020image} & 6/1.3 & 72.5 & 71.1 & 7.5 & 33.0 & 20.1 \\
    \rowcolor{Gray}
    - AbS & 7/2.6 & \textbf{74.1} & \textbf{66.7} & \textbf{10.1} & \textbf{34.9} & \textbf{22.6} \\
    \midrule
    RVT-Ti~\cite{mao2022towards} & 9/1.3 & 78.1 & 58.8 & 13.9 & 42.5 & 29.1 \\
    \rowcolor{Gray}
    - AbS & 11/2.7 & \textbf{78.6} & \textbf{55.9} & \textbf{17.3} & \textbf{43.2} & \textbf{29.9} \\
    \midrule
    FAN-Ti~\cite{zhou2022understanding} & 7/1.3 & 77.5 & 59.8 & 13.1 & 42.6 & 29.9 \\
    \rowcolor{Gray}
    - AbS & 9/2.9 & \textbf{78.3} & \textbf{57.4} & \textbf{16.5} & \textbf{42.8} & \textbf{31.2} \\
    \midrule

    PiT-S~\cite{heo2021rethinking} & 24/2.9 & 80.9 & 52.5 & 21.7 & 43.6 & 30.8 \\
    PVT-S~\cite{wang2021pyramid} & 25/3.8 & 79.9 & 66.9 & 18.0 & 40.1 & 27.2 \\
    Swin-T~\cite{liu2021swin} & 28/4.5 & 81.2 & 62.0 & 21.6 & 41.3 & 29.1 \\
    ConvNext-T~\cite{liu2022convnet} & 29/4.5 & 82.1 & 53.2 & 24.2 & 47.2 & 33.8 \\
    \midrule
    ViT-S~\cite{dosovitskiy2020image} & 22/4.2 & 80.1 & 54.6 & 19.2 & 41.9 & 28.9 \\
    \rowcolor{Gray}
    - AbS & 26/9.8 & \textbf{80.7} & \textbf{51.6} & \textbf{24.3} & \textbf{43.1} & \textbf{30.2} \\
    \midrule
    RVT-S~\cite{mao2022towards} & 22/4.3 & 81.9 & 50.5 & 26.0 & 47.0 & 34.5 \\
    \rowcolor{Gray}
    - AbS & 26/10.4 & 81.9 & \textbf{48.7} & \textbf{31.1} & \textbf{48.5} & \textbf{35.6} \\
    \midrule
    FAN-S~\cite{zhou2022understanding} & 28/5.3 & 82.8 & 49.1 & 29.3 & 47.4 & 35.6 \\
    \rowcolor{Gray}
    - AbS & 32/11.4 & \textbf{83.0} & \textbf{47.4} & \textbf{34.0} & \textbf{48.3} & \textbf{36.4}\\
    \midrule

    PiT-B~\cite{heo2021rethinking} & 74/12.5 & 82.4 & 48.2 & 33.9 & 43.7 & 32.3 \\
    PVT-L~\cite{wang2021pyramid} & 61/9.8 & 81.7 & 59.8 & 26.6 & 42.7 & 30.2 \\
    Swin-B~\cite{liu2021swin} & 88/15.4 & 83.4 & 54.4 & 35.8 & 46.6 & 32.4 \\
    ConvNext-B~\cite{liu2022convnet} & 89/15.4 & 83.8 & 46.8 & 36.7 & 51.3 & 38.2 \\
    \midrule
    ViT-B~\cite{dosovitskiy2020image} & 87/17.2 & 80.8 & 49.3 & 25.2 & \textbf{43.3} & 31.6 \\
    \rowcolor{Gray}
    - AbS & 99/38.9 & \textbf{81.0} & \textbf{48.3} & \textbf{28.2} & 42.9 & 31.7 \\
    \midrule
    RVT-B~\cite{mao2022towards} & 86/17.7 & 80.9 & 52.1 & 26.6 & \textbf{39.6} & 26.1 \\
    \rowcolor{Gray}
    - AbS & 100/39.5 & 80.9 & \textbf{51.7} & \textbf{28.5} & 39.3 & 26.0 \\
    \midrule
    FAN-B~\cite{zhou2022understanding} & 54/10.4 & 83.5 & 45.0 & 33.2 & 51.4 & 39.3 \\
    \rowcolor{Gray}
    - AbS & 62/21.8 & \textbf{83.7} & \textbf{44.1} & \textbf{38.4} & \textbf{52.0} & \textbf{39.8} \\
    
    \bottomrule
  \end{tabular}
  \caption{Results on ImageNet classification and robustness benchmarks. \model improves performance across different benchmarks and backbones. P/F: \# of parameters and FLOPs. $\downarrow$: lower is better.}
  \label{tab:imagenet}
  \end{scriptsize}
\end{table}

\begin{table}[t]
  \centering
  \begin{footnotesize}
  \begin{tabular}{@{}lccccc@{}}
    Model & Clean & IN-C ($\downarrow$) & IN-A & IN-R & IN-SK \\
    \midrule    
    ViT-Ti & 72.5 & 71.1 & 7.5 & 33.0 & 20.1 \\
    - PerceiverIO & 72.8 & 70.4 & 8.0 & 32.8 & 20.5 \\
    - Feedback & 73.4 & 67.8 & 9.7 & 34.6 & 22.4 \\
    - MaskAtt & 72.5 & 70.6 & 8.3 & 33.4 & 20.5 \\
    - AbS & \textbf{74.1} & \textbf{66.7} & \textbf{10.1} & \textbf{34.9} & \textbf{22.6} \\   
    \midrule
    RVT-Ti & 78.1 & 58.8 & 13.9 & 42.5 & 29.1 \\
    - PerceiverIO & 78.3 & 57.8 & 13.7 & 42.8 & 29.8 \\
    - Feedback & 79.1 & 55.7 & 18.2 & 44.1 & 31.3 \\
    - MaskAtt & 77.9 & 59.0 & 13.5 & 43.0 & 29.7 \\
    - AbS & \textbf{79.5} & \textbf{54.8} & \textbf{18.7} & \textbf{44.5} & \textbf{32.5} \\
\bottomrule
  \end{tabular}
  \caption{Comparison of different top-down attention algorithms on ImageNet classification and robustness.}
  \label{tab:ablation_att}
  \end{footnotesize}
\end{table}

\begin{figure}[t]
\begin{center}
\centerline{\includegraphics[width=0.9\columnwidth]{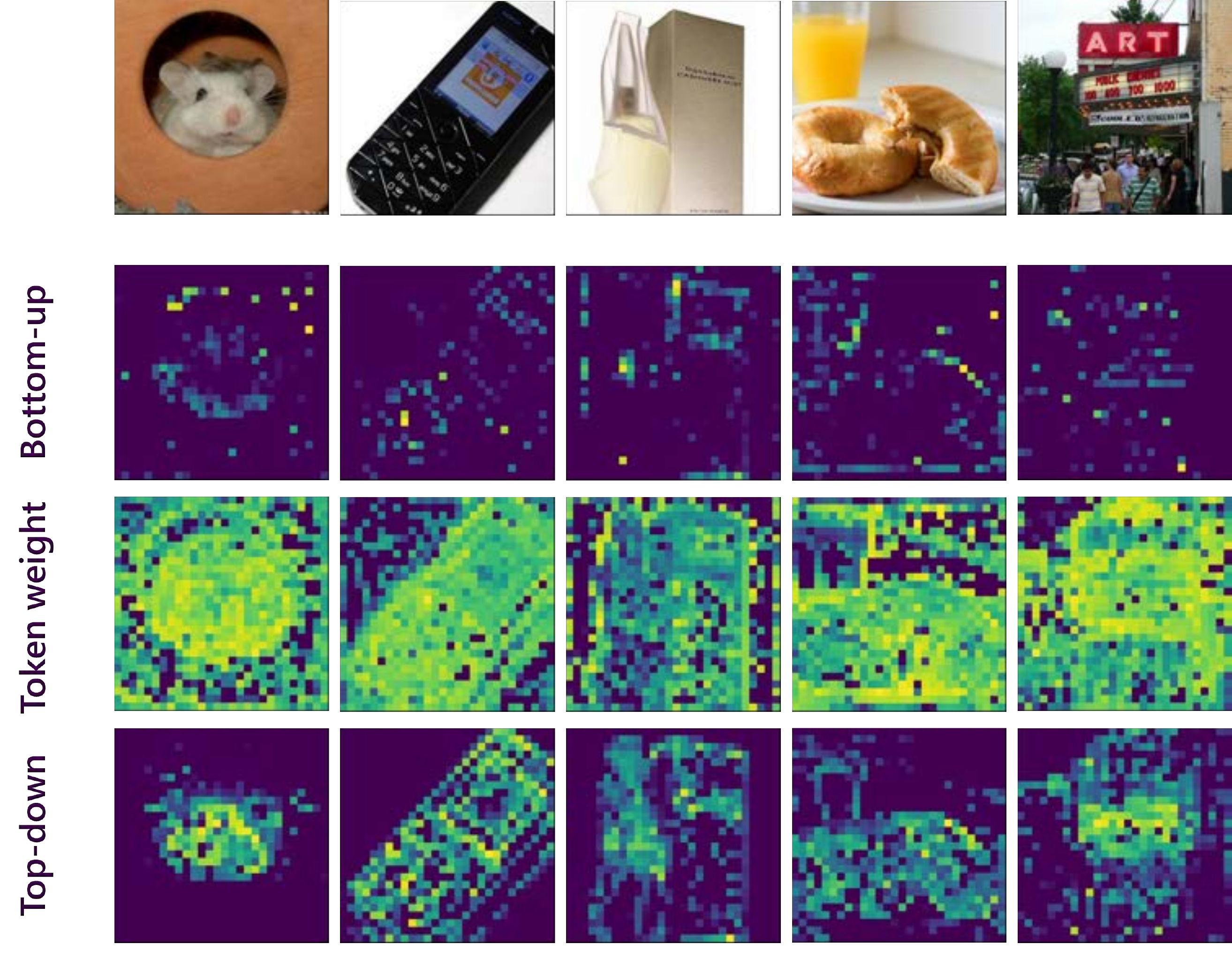}}
\caption{Visualization of the bottom-up attention, token weights, and the top-down attention in \model. The bottom-up attention is noisy and fails to detect the complete foreground object. In \model, the query mask can coarsely detect the foreground object and reweight tokens fed back to direct the top-down attention to better extract the foreground object. }
\label{fig:image_classification}
\end{center}
\vskip -0.3in
\end{figure}

We test \model on ImageNet classification and robustness benchmarks (\cref{tab:imagenet}). %
We report mCE (lower the better)~\cite{hendrycks2019benchmarking} for IN-C and accuracy for other datasets. On clean images, AbSViT consistently improves over baselines, with a similar number of parameters although higher FLOPs. The clean accuracy on FAN-B is improved to 83.7\%, reaching the same level as ConvNext-B with fewer parameters. On corrupted (IN-C) and adversarial (IN-A) images, \model boosts the performance by about $1$-$5\%$ across all the scales. Especially, the performance on FAN-B is raised by $1\%$ and $5\%$ for IN-C and IN-A, reaching a new state-of-the-art result. On out-of-distribution images, \model also improves by $3\%$ on Tiny and Small models and $0.5\%$ on FAN-B.

\cref{fig:image_classification} visualizes the attention map of ViT and \model, as well as token weights generated in $e_\xi(\cdot)$. The bottom-up attention in ViT is often noisy and only partly detects the foreground object. On the other hand, the query $\xi$ in \model learns to coarsely detect the foreground and reweight the feedforward output tokens, which are fed back and generate top-down attention that better detects the foreground object.

We compare \model with several baseline algorithms for goal-directed attention in \cref{tab:ablation_att}. One may see that a pure feedback model already improves the clean accuracy and robustness, and \model further boosts the performance by better extracting the foreground object. Due to a similar reason, PerceiverIO without feedback also slightly improves the performance. On the other hand, MaskAtt is sometimes harmful (on Clean, IN-C, and IN-A for RVT), implying that a mask attention design is unsuitable for vision transformers.

\subsection{Semantic Segmentation}
\label{sec:exp_segmentation}

\begin{table}[t]
  \centering
  \begin{footnotesize}
  \begin{tabular}{@{}lccc@{}}
    Model & PASCAL VOC & Cityscapes & ADE20K \\
    \midrule
    ResNet-101~\cite{mmseg2020} & 77.1 & \textbf{78.7} & 42.9 \\
    \midrule    
     ViT-B & 80.1 & 75.3 & 45.2 \\
     AbSViT-B & \textbf{81.3} \scriptsize(\textcolor{LimeGreen}{+1.2}) & 76.8 \scriptsize(\textcolor{LimeGreen}{+1.5}) &  \textbf{47.2} \scriptsize(\textcolor{LimeGreen}{+2.0}) \\
     \bottomrule
  \end{tabular}
  \caption{Semantic segmentation results on three datasets.}
  \label{tab:seg}
  \end{footnotesize}
  \vskip -0.1in
\end{table}

We evaluate the performance of \model as a backbone for semantic segmentation on three datasets (PASCAL VOC, Cityscapes, and ADE20K). We compare with two baseline backbones, regular ViT and ResNet-101. We use UperNet~\cite{xiao2018unified} as the segmentation head for all the backbones. Results are shown in \cref{tab:seg}. We can see that when using \model as the backbone, we can achieve $1.2$-$2.0\%$ improvements over the ViT baseline with approximately the same number of parameters. This indicates that \model can be used as a general backbone for different vision tasks.

\subsection{Justification of Model Design}
\label{sec:exp_model_design}

The design of \model follows the principle of AbS. For example, \model adds the top-down signal only to the value matrix considering the analogy between self-attention and sparse reconstruction (\cref{sec:design}). At the same time, an arbitrary design may also add it to the query and key. We also optimize the variational loss to approximate AbS instead of just building a top-down model and training with the supervised loss. In this section, we show the advantage of these ``destined'' designs compared with an arbitrary design, which also justifies the proposed guiding principle of AbS.

We first try an arbitrary design of self-attention with top-down input by adding the top-down signal on the query, key, and value instead of only on the value. We name this design as \model-QKV. We compare \model and \model-QKV on image classification and robustness (\cref{tab:ablation_qkv}), and we can see that \model is superior to \model-QKV on every benchmark. This is consistent with our analysis in \cref{sec:td_from_abs} that the sparse reconstruction AbS is optimizing has an additional top-down input (corresponding to V), while the dictionary (corresponding to Q and K), which contains templates for separate objects, is fixed. 

We also test the effect of the variational loss $\mathcal{L}_{var}$, which ensures the model is approximating AbS. We compare \model with its counterpart without $\mathcal{L}_{var}$, \ie, a top-down model trained with only supervised loss. As shown in \cref{tab:ablation_var}, adding $\mathcal{L}_{var}$ largely improves the clean accuracy and robustness. Note that, as discussed in \cref{sec:design}, we do not have a prior loss $-\log p(\bfz_L)$ for image classification, which means the improvement completely comes from the reconstruction loss $\frac{1}{2} \sum_{\ell=1}^L ||\mathit{sg}(\bfz_\ell) - \mathit{g}_\ell(\mathit{sg}(\bfz_{\ell+1}))||_2^2$ which forces the decoder to reconstruct $\bfz_\ell$ from $\bfz_{\ell+1}$. This implies that a generative model (``synthesis'') is important to high-quality top-down attention in visual recognition (``analysis'').

\begin{table}[t]
  \centering
  \begin{footnotesize}
  \begin{tabular}{@{}lccccc@{}}
    Model & Clean & IN-C ($\downarrow$) & IN-A & IN-R & IN-SK \\
    \midrule    
    \model-QKV & 73.3 & 68.0 & 9.4 & 33.8 & 21.2 \\
    \model & \textbf{74.1} & \textbf{66.7} & \textbf{10.1} & \textbf{34.9} & \textbf{22.6} \\   
\bottomrule
  \end{tabular}
  \caption{The predicted design of top-down self-attention (\model) is better than an arbitrary design (\model-QKV).}
  \label{tab:ablation_qkv}
  \end{footnotesize}
\end{table}

\begin{table}[t]
  \centering
  \begin{footnotesize}
  \begin{tabular}{@{}lcccccc@{}}
     & $\mathcal{L}_{var}$ & Clean & IN-C ($\downarrow$) & IN-A & IN-R & IN-SK \\
    \midrule    
    \model & \xmark & 73.1 & 69.0 & 9.5 & 33.5 & 20.8 \\
    \model & \cmark & \textbf{74.1} & \textbf{66.7} & \textbf{10.1} & \textbf{34.9} & \textbf{22.6} \\   
\bottomrule
  \end{tabular}
  \caption{Ablation on the variational loss $\mathcal{L}_{var}$.}
  \label{tab:ablation_var}
  \end{footnotesize}
  \vskip -0.1in
\end{table}
\section{Limitations and Future Work}

\subsection{ImageNet Classification Is a Poor Teacher of Top-Down Attention}
\label{sec:lim_weak_td}

\begin{figure}[t]
\begin{center}
\centerline{\includegraphics[width=0.9\columnwidth]{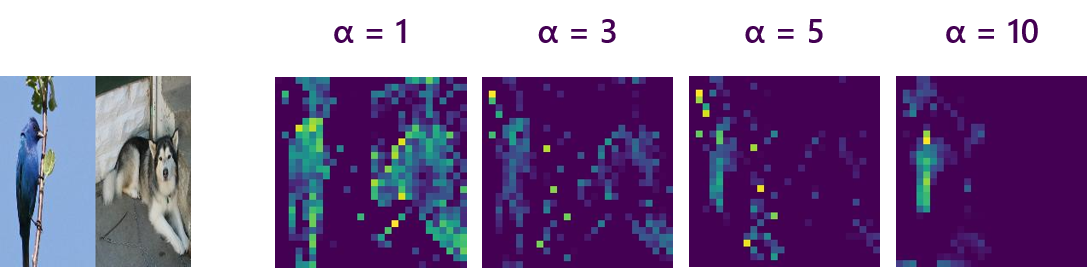}}
\caption{Visualization of top-down attention with different scaling factor $\alpha$. Prior corresponds to the bird. The top-down attention gets more and more biased on the bird when increasing $\alpha$.}
\label{fig:weak_attention}
\end{center}
\vskip -0.3in
\end{figure}

\model is trained to focus on different objects given different priors in multi-object images. However, ImageNet classification targets single object classification without any prior, making it unsuitable for pretraining top-down attention.
We find that the ImageNet-supervised \model only learns weak top-down attention. A simple trick to augment the top-down attention for downstream tasks such as VQA is manually setting a larger scaling factor $\alpha$ (\eg, $\alpha = 10$). In \cref{fig:weak_attention}, we visualize the top-down attention with different $\alpha$. We can see that, with a prior corresponding to the bird, the attention under $\alpha=1$ still highlights both the bird and the dog but is more and more biased towards the bird as we increase $\alpha$. For future exploration, we may learn stronger top-down attention through object-level unsupervised learning~\cite{xiao2021region,henaff2022object} or vision-language pretraining~\cite{xu2022groupvit,mukhoti2022open}.

\subsection{How Many Syntheses Do We Need for Analysis?}
\label{sec:lim_weak_syn}

\begin{figure}[t]
\begin{center}
\centerline{\includegraphics[width=0.9\columnwidth]{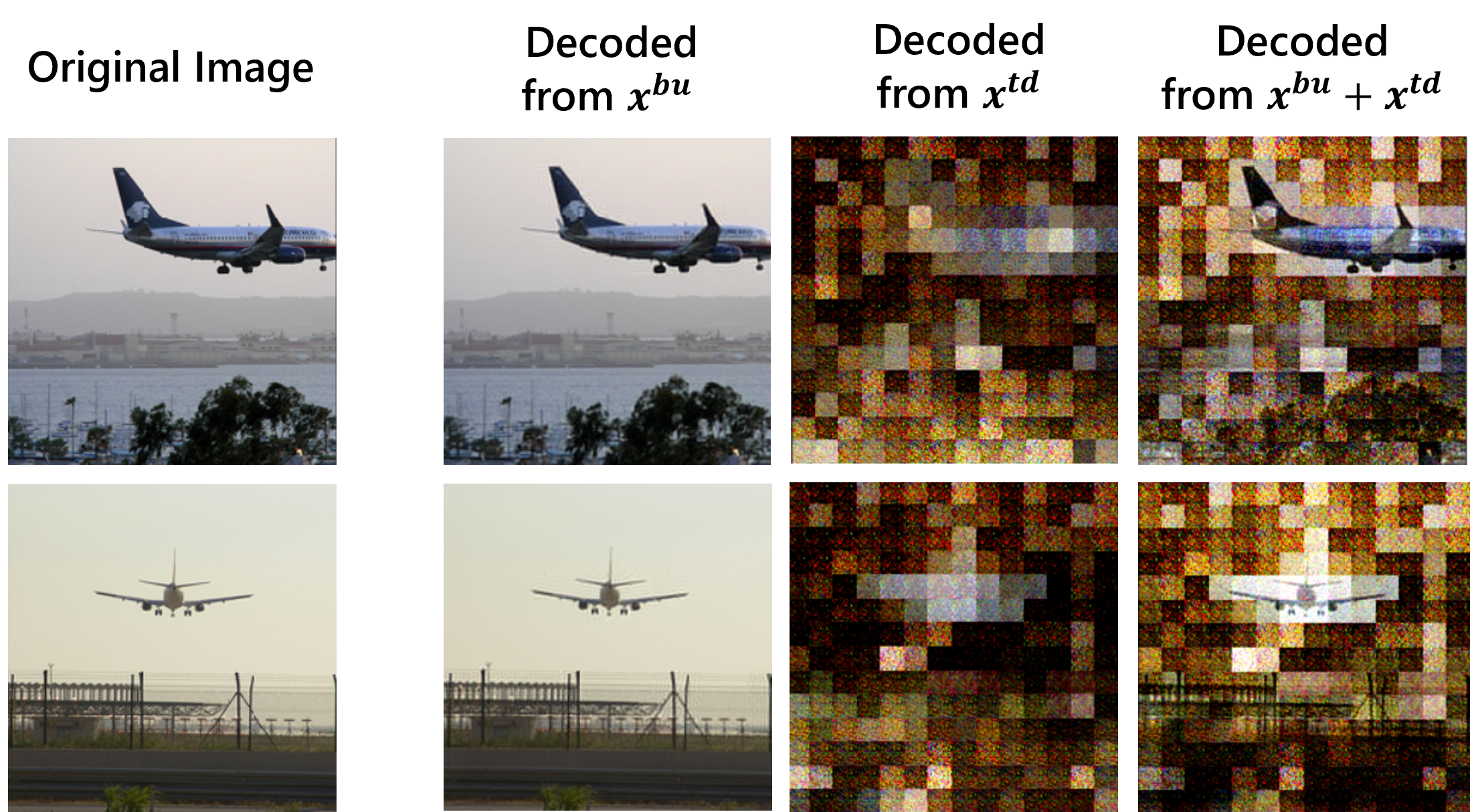}}
\caption{Examples of images decoded from the bottom-up, top-down, or the combination of bottom-up and top-down signals. The decoder can reconstruct the whole image from the bottom-up signal while failing to generate anything recognizable from the top-down signal alone. When decoding from the combination of bottom-up and top-down signals, only the foreground object is reconstructed.   }
\label{fig:weak_synthesis}
\end{center}
\vskip -0.4in
\end{figure}

In \cref{sec:implem}, we mention that enforcing strong generative capability on the features $\bfz_\ell$ will downgrade the discriminative power regarding classification accuracy. There is a similar observation in recent self-supervised learning work~\cite{he2022masked}, where reconstruction-based algorithms have worse linear-probing performance~\cite{caron2021emerging}. However, the empirical results in \cref{tab:ablation_var} indicate that at least some degree of generative power is still helpful. This echoes the classical debate of how much generative capability (``synthesis'') we need for visual discrimination (``analysis''). As a starting point, we measure the generative power of the ImageNet-pretrained \model (\cref{fig:weak_synthesis}). Specifically, we train a linear decoder that projects the bottom-up input $\bfx_0^{bu}$ of the first layer to the original image and then visualize the image decoded from the bottom-up signal $\bfx_0^{bu}$, the top-down signal $\bfx_0^{td}$, or their combination $\bfx_0^{bu} + \bfx_0^{td}$. We can see that the bottom-up signal contains full information about the original image and gives a perfect reconstruction. On the other hand, the top-down signal has lost most of the information, which is reasonable considering that $\bfx_0^{td}$ itself is decoded from the last layer's feature. Intriguingly, when we combine the bottom-up and the top-down signals, it can reconstruct only the foreground object, implying \model can selectively preserve partial information in the image, and the selection process is adaptive to different priors. This leaves the question of whether a \textit{selective} generation process is the best companion of the discriminative model and how to control the selective process under different priors adaptively.
\section{Conclusion}
\label{sec:conclusion}

We consider top-down attention by explaining from an Analysis-by-Synthesis (AbS) view of vision. Starting from previous work on the functional equivalence between visual attention and sparse reconstruction, we show that AbS optimizes a similar sparse reconstruction objective but modulates it with a goal-directed top-down modulation, thus simulating top-down attention. We propose \model, a top-down modulated ViT model that variationally approximates AbS. We show that \model achieves controllable top-down attention and improves over baselines on V\&L tasks as well as image classification and robustness.

\minisection{Acknowledgement}.
The authors would like to thank Tianyuan Zhang and Amir Bar for their valuable suggestions. Baifeng Shi and Trevor Darrell are supported by DARPA and/or the BAIR Commons program.

{\small
\bibliographystyle{plainnat}
\bibliography{egbib}
}

\newpage
\appendix
\onecolumn

\section{Derivation of Eq. (10)}

From Eq. (6-7) we have
\begin{equation}\small
    p(\tbfu_\ell | \tbfu_{\ell+1}) \propto \exp\{-\frac{1}{2}||\bfP_\ell \tbfu_\ell - \mathit{g}_\ell(\bfP_{\ell+1} \tbfu_{\ell+1})||_2^2 - \lambda||\tbfu_\ell||_1\}.
\end{equation}
Then Eq. (10) is derived by
\begin{equation}\small
\begin{split}
    \dv{\tbfu_\ell}{t} &\propto \nabla_{\tbfu_\ell} \log p(\tbfu_{\ell-1} | \tbfu\ell) + \nabla_{\tbfu_\ell} \log p(\tbfu_\ell | \tbfu_{\ell-1}) \\
    &= -\nabla_{\tbfu_\ell}\frac{1}{2} ||\bfP_{\ell-1} \tbfu_{\ell-1} - \mathit{g}_{\ell-1} (\bfP_\ell \tbfu_\ell)||_2^2 -\nabla_{\tbfu_\ell}\frac{1}{2} ||\bfP_\ell \tbfu_\ell - \mathit{g}_\ell (\bfP_{\ell+1} \tbfu_{\ell+1})||_2^2 - \nabla_{\tbfu_\ell} \lambda ||\tbfu_\ell||_1 \\
    &= \bfP_\ell^T \bfJ_{\ell-1}^T \left(\bfP_{\ell-1} \tbfu_{\ell-1} - \mathit{g}_{\ell-1} (\bfP_\ell \tbfu_\ell)\right) - \bfP_\ell^T \left(\bfP_\ell \tbfu_\ell - \mathit{g}_\ell (\bfP_{\ell+1} \tbfu_{\ell+1}) \right) - \nabla_{\tbfu_\ell} \lambda ||\tbfu_\ell||_1\\
    &= - \bfP_\ell^T \left(\bfP_\ell \tbfu_\ell - \mathit{g}_\ell (\bfP_{\ell+1} \tbfu_{\ell+1}) - \bfJ_{\ell-1}^T \bfP_{\ell-1} \tbfu_{\ell-1} \right) - \nabla_{\tbfu_\ell} \lambda ||\tbfu_\ell||_1 - \bfP_\ell^T \bfJ_{\ell-1}^T \mathit{g}_{\ell-1} (\bfP_\ell \tbfu_\ell)\\
    &= -\nabla_{\tbfu_\ell}\frac{1}{2} ||\bfP_\ell \tbfu_\ell - \mathit{g}_\ell (\bfz_{\ell+1}) - \bfJ_{\ell-1}^T \bfz_{\ell-1}||_2^2 - \nabla_{\tbfu_\ell} \lambda ||\tbfu_\ell||_1 - \nabla_{\tbfu_\ell} \frac{1}{2} ||\mathit{g}_{\ell-1} (\bfP_\ell \tbfu_\ell)||^2_2 \\
    &= -\nabla_{\tbfu_\ell}\frac{1}{2} ||\bfP_\ell \tbfu_\ell - (\bfx_\ell^{td} + \bfx_\ell^{bu})||_2^2 - \nabla_{\tbfu_\ell} \lambda ||\tbfu_\ell||_1 - \nabla_{\tbfu_\ell} \frac{1}{2} ||\mathit{g}_{\ell-1} (\bfP_\ell \tbfu_\ell)||^2_2.
\end{split}
\end{equation}
We informally use $\nabla$ for subgradients as well.

\section{Additional Results on Natural Images}

\begin{figure}[h]
\begin{center}
\centerline{\includegraphics[width=0.45\columnwidth]{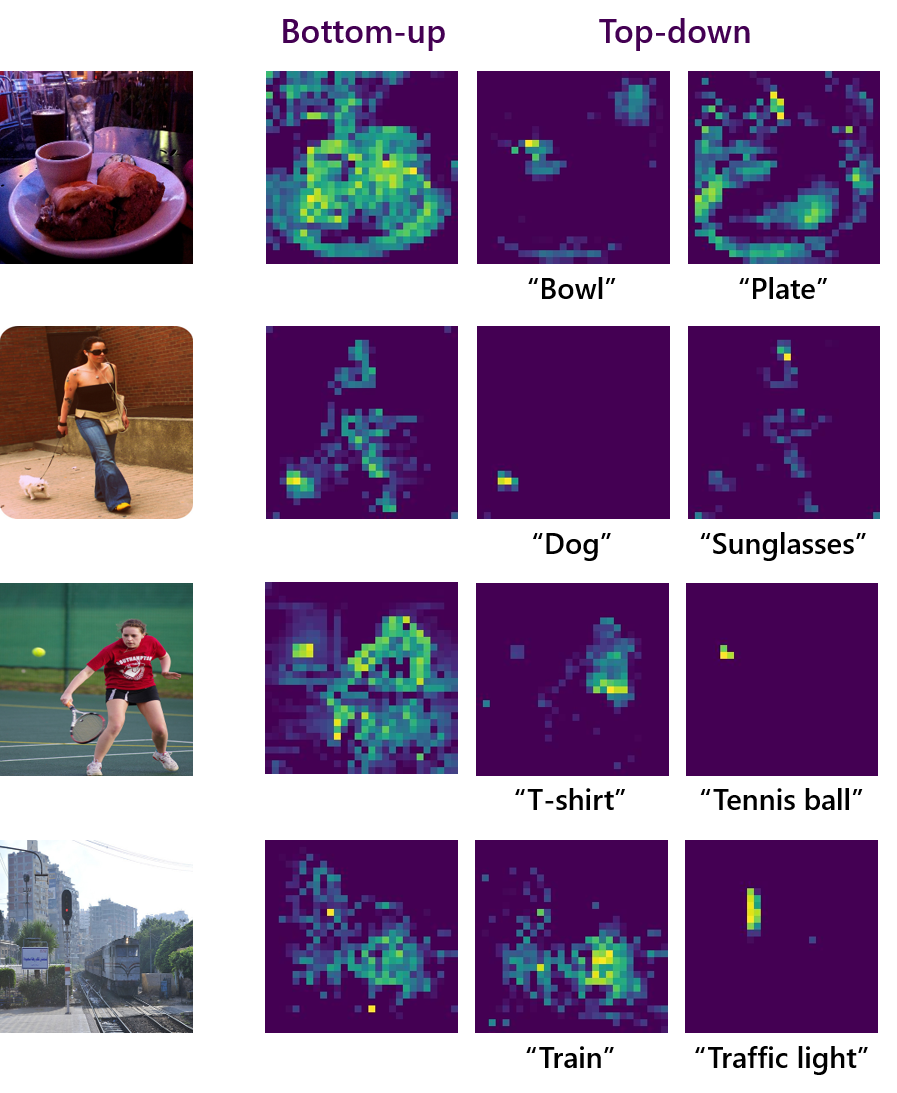}}
\caption{Visualization of top-down attention on natural images. From left to right, we show the original images, the bottom-up attention, as well as the top-down attention regarding to different objects in each image.}
\label{fig:natural_spatial_bistable}
\end{center}
\vskip -0.3in
\end{figure}

\begin{figure}[H]
\begin{center}
\centerline{\includegraphics[width=0.5\columnwidth]{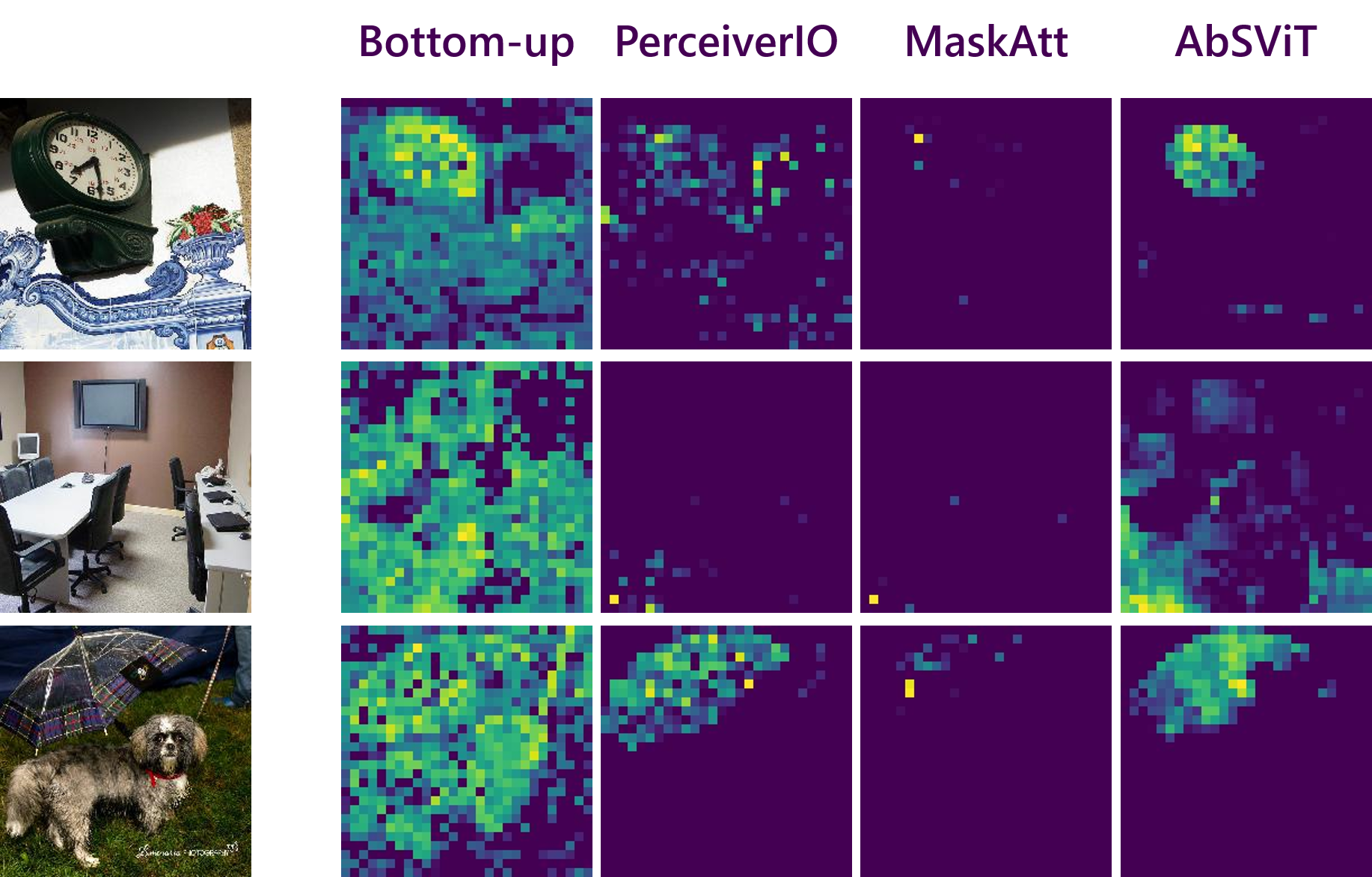}}
\caption{Comparison of top-down attention map between \model and different baselines. }
\label{fig:natural_compare}
\end{center}
\end{figure}

In Fig.\ref{fig:spatial_bistable}-\ref{fig:att_compare}, we show examples of top-down attention on artificial images. Here we show more results on natural images containing multiple objects. We borrow the LVIS dataset and collect images that contain object categories that also appear in ImageNet. We demonstrate that given different prior, \model is able to focus on different objects in the same image (\cref{fig:natural_spatial_bistable}). We also compare \model's top-down attention with several baseline methods (\cref{fig:natural_compare}) and observe that \model has cleaner attention maps than other methods.

\section{Additional Implementation Details}

\minisection{ImageNet Pretraining}. The ViT and RVT baselines as well as our \model model are trained using the recipe in \cite{mao2022towards}, and FAN is trained using the recipe in its original paper~\cite{zhou2022understanding}. Specifically, we use AdamW optimizer to train \model for 300 epochs, with a batch size of 512, a base learning rate of 5e-4, and 5 warm-up epochs. One may use different batch-size and adjust the learning rate by the linear scaling rule. We use a cosine learning rate scheduling and weight decay of 0.05. We use the default setting of data augmentation, which includes Mixup, Cutmix, ColorJittering, AutoAugmentation, and Random Erasing. For \model, the weights of supervised loss and variational loss are set as 1 and 0.1.

\minisection{Robustness against Image Corruptions}. We evaluate model robustness against image corruption on ImageNet-C, which contains a total of 19 corruption types. We follow \cite{mao2022towards} and evaluate 15 types of corruption including Brightness, Contrast, Defocus Blur, Elastic Transform, Fog, Frost, Gaussian Noise, Glass Blur, Impulse Noise, JPEG Compression, Motion Blur, Pixelate, Shot Noise, Snow, and Zoom Blur. Note that other work (e.g. \cite{zhou2022understanding}) tests on a different subset of corruption types. To make a fair comparison, all the models are tested under the aforementioned 15 corruption types.

\minisection{Semantic Segmentation}. We use MMSegmentation~\cite{mmseg2020} as our test bed. We take the ImageNet pretrained ViT-B and \model-B and finetune them on semantic segmentation on PASCAL VOC, Cityscapes, and ADE20K. For all the experiments, we use UperNet~\cite{xiao2018unified} as the decoder head and FCNHead as the auxiliary head. We train on 2 GPUs with a total batch size of 16, using AdamW optimizer, a learning rate of $0.00006$, and weight decay of $0.01$. We train for 20k, 40k, and 160k iterations for three datasets, respectively. We use image resolution of 512x512 for PASCAL VOC and ADE20K, and 512x1024 for Cityscapes.

\minisection{V\&L Finetuning}. Following \cite{dou2022empirical}, the whole model contains a pretrained visual encoder, a pretrained text encoder, and a multimodal encoder to merge vision and language. We use the ImageNet pretrained ViT or \model for the visual encoder, a pretrained RoBERTa for the text encoder, and the multimodal encoder is trained from scratch.  We use a learning rate of $1e-5$ for visual and text encoders and $5e-5$ for the multimodal encoder. For top-down attention, we use the \texttt{[cls]} token as the prior $\xi$. Since the text and visual tokens are not aligned initially, we train a linear transform to project the text tokens into the same space as the visual tokens. This is trained by the prior loss, which is set as a CLIP-style loss (\cref{eq:clip_loss}) to align the text and visual tokens.

\end{document}